\documentclass[11pt]{article}

\usepackage{amsmath}
\usepackage{amssymb}
\usepackage{booktabs}
\usepackage{multirow}
\usepackage[most]{tcolorbox}
\usepackage{marvosym}
\tcbuselibrary{listings, breakable}
\usepackage[final]{acl}

\usepackage{times}
\usepackage{latexsym}
\usepackage{xcolor}

\usepackage[T1]{fontenc}

\usepackage[utf8]{inputenc}

\usepackage{microtype}

\usepackage{inconsolata}

\usepackage{graphicx}

\title{SimRPD: Optimizing Recruitment Proactive Dialogue Agents through Simulator-Based Data Evaluation and Selection}

\author{
 \textbf{Zhiyong Cao\textsuperscript{1,2\footnotemark[1]}},
 \textbf{Dunqiang Liu\textsuperscript{1,3\footnotemark[1]}},
 \textbf{Qi Dai\textsuperscript{1,4\footnotemark[1]}},
 \textbf{Haojun Xu\textsuperscript{1,5\footnotemark[1]}},
 \textbf{Huai Yuen Khor\textsuperscript{1\footnotemark[1]}},
\\
\textbf{Hao Wang\textsuperscript{1}},
 \textbf{Huan He\textsuperscript{1}},
 \textbf{Yafei Liu\textsuperscript{1}},
 \textbf{Ke Ma\textsuperscript{1}},
 \textbf{Ruqian Shi\textsuperscript{6}},
   \textbf{Sicheng Zhou\textsuperscript{1\footnotemark[2]}},
 \textbf{Sijia Yao\textsuperscript{1\footnotemark[2]}}
\\
\\
 \textsuperscript{1}Zhaopin Limited \quad
 \textsuperscript{2}Institute of Automation, Chinese Academy of Sciences
 \\
 \textsuperscript{3}School of Automation,Beijing Institute of Technology,
 \\
 \textsuperscript{4}Central University of Finance and Economics \quad
 \textsuperscript{5}Beihang University \quad
 \textsuperscript{6}Independent Researcher
 \\
 \small{
   \textbf{Correspondence:} 
 \href{zhou.sicheng@zhaopin.com.cn}{zhou.sicheng@zhaopin.com.cn},
\href{yaosj07@gmail.com}{yaosj07@gmail.com}
}
}

\begin{document}
\maketitle

\begingroup
\renewcommand{\thefootnote}{\fnsymbol{footnote}}
\footnotetext[1]{Equal contribution.}
\footnotetext[2]{Corresponding author.}
\endgroup

\begin{abstract}
Task-oriented proactive dialogue agents play a pivotal role in recruitment, particularly for steering conversations towards specific business outcomes, such as acquiring social-media contacts for private-channel conversion. Although supervised fine-tuning and reinforcement learning have proven effective for training such agents, their performance is heavily constrained by the scarcity of high-quality, goal-oriented domain-specific training data. To address this challenge, we propose \textbf{SimRPD}, a three-stage framework for training recruitment proactive dialogue agents. First, we develop a high-fidelity user simulator to synthesize large-scale conversational data through multi-turn online dialogue. Then we introduce a multi-dimensional evaluation framework based on \textbf{Chain-of-Intention (CoI)} to comprehensively assess the simulator and effectively select high-quality data, incorporating both global-level and instance-level metrics. Finally, we train the recruitment proactive dialogue agent on the selected dataset. Experiments in a real-world recruitment scenario demonstrate that SimRPD outperforms existing simulator-based data selection strategies, highlighting its practical value for industrial deployment and its potential applicability to other business-oriented dialogue scenarios.
\end{abstract}

\section{Introduction}

The advent of Large Language Models (LLMs) \citep{guo2025deepseek, achiam2023gpt} has reshaped Task-Oriented Dialogue (TOD) systems \citep{kwan2023survey, qin2023end}, transitioning them from rigid slot-filling pipelines to versatile agents capable of multi-step reasoning and strategy \citep{ouyang2022training, zhao2023survey}. In high-stakes domains like recruitment, TOD systems are increasingly developed as Proactive Dialogue (PD) agents \citep{chouhan2025answers, zhang2025enhancing}. Unlike traditional customer service agents that respond to customers' questions passively, these agents must actively steer multi-turn conversations to achieve measurable business outcomes \citep{wang2024target, wang2022follow}. Specifically, Recruitment Proactive Dialogue (RPD) agents are designed to interact with candidates over multiple turns and complete a private-channel handoff (e.g. persuading the candidate to click a contact card) for downstream conversion.

\begin{figure}[t]
    \centering
     \includegraphics[width=\linewidth]{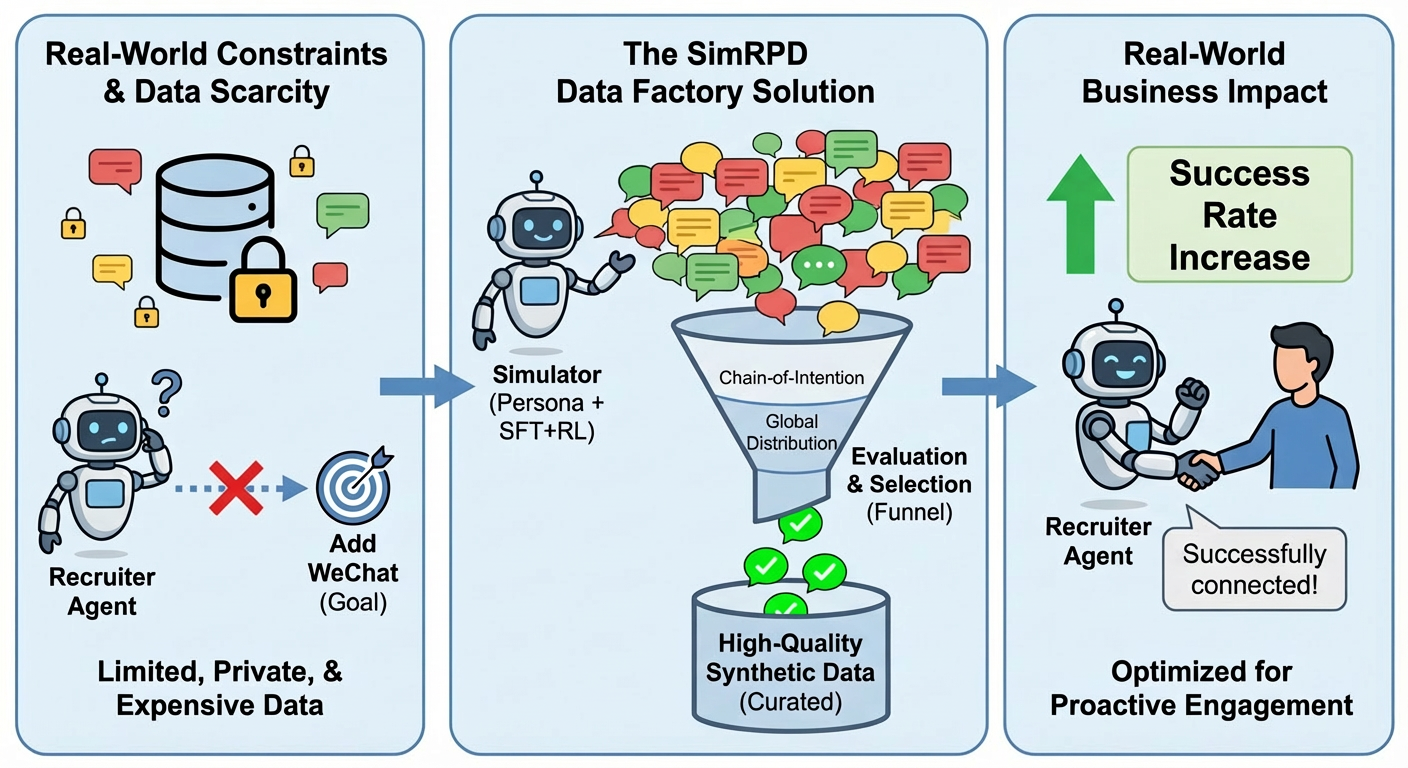}
  \caption{Background of this work. High-quality data in training proactive dialogue agents in real-world applications is sparse, therefore we train a user simulator to synthesize data and propose a dual-level evaluation protocol to select premium data.}
  \label{fig:background}
\end{figure}

Existing research on PD agents has mainly focused on knowledge retrieval \citep{dong2025protod}, prompt engineering \citep{wang2023target}, and supervised fine-tuning \citep{li2025two}. However, training such agents is bottlenecked by severe \textbf{data sparsity} (Figure~\ref{fig:background}). Existing methods are evaluated in simple domains such as daily dialogue or commodity recommendation, while high-quality, goal-oriented interaction data on complex scenarios like job consulting is lacking, where dialogue models combine rich domain knowledge and can understand user profile and intent. Consequently, Synthetic Data Generation (SDG) via user simulators has emerged as a promising solution. Recent works such as CAMEL \citep{li2023camel} and Generative Agents \citep{park2023generative} have demonstrated that LLMs can act as high-fidelity simulators to produce vast amounts of interaction data, effectively creating a sandbox for training downstream models without exposing real users to experimental risks.

Despite the potential of SDG, there are two main challenges. \textbf{(1) High-Quality Data Generation}. While it is easy to generate millions of dialogue turns using LLMs, recent studies warn that blindly training on synthetic data can lead to model collapse or the amplification of biases \citep{shumailov2023curse, gudibande2023false}. Most existing simulators suffer from sycophancy (being overly agreeable) or lack the diverse behavioral distributions found in real job seekers \citep{sharma2023towards}, leading to downstream agents that fail when facing real-world rejections. \textbf{(2) Comprehensive Evaluation}. Prevailing evaluation metrics for simulators focus on instance-level metrics like fluency or coherence based on LLMs \citep{liu2023g, wang2025know}. However, they overlook global-level metrics such as synthetic data distribution with authentic data or diversity of user queries. Therefore, we need a multi-dimensional evaluation framework that can assess simulator fidelity and support reliable data filtering.

To bridge these gaps, we introduce SimRPD, a framework for training recruitment proactive dialogue agents via simulator-driven data generation and rigorous selection. Inspired by \citep{wang2025know}, we train a user simulator with Supervised Fine-Tuning (SFT) and Reinforcement Learning (RL) to mimic the nuanced resistance and negotiation patterns of real candidates. We further propose a multi-dimensional evaluation protocol based on Chain-of-Intention (CoI) that assesses synthetic data at both the instance-level and global-level. This evaluation protocol serves two purposes: evaluating user simulator fidelity and selecting high-quality synthetic data, thereby filtering out low-value or misleading samples that could mislead the agent. Finally, we train the recruitment PD agent on the selected data via SFT and RL, which is then deployed in a real-world recruitment platform.

In summary, our contributions are as follows:

\begin{itemize}
    \item We propose SimRPD, a framework that integrates user simulation with a rigorous data selection mechanism and mitigates data sparsity and privacy constraints in training recruitment proactive dialogue agents.
    \item We introduce a dual-level evaluation protocol that jointly assesses simulator fidelity and filters synthetic dialogues, improving both instance-level quality and global distributional alignment with real interactions.
    \item We demonstrate the effectiveness of our approach through large-scale industrial deployment, showing tangible gains in proactive business goals.
\end{itemize}

\section{Related Works}

\subsection{Task-Oriented Proactive Dialogue Systems}
Task-Oriented Dialogue (TOD) systems have evolved from passive query-responding models to proactive agents. Early approaches relied on pipeline architectures optimized by Reinforcement Learning (RL) for slot-filling tasks \citep{young2013pomdp,williams2007partially}. With the advent of Large Language Models, research has shifted towards end-to-end agents capable of negotiation, persuasion, and target-oriented transitions \citep{lewis2017deal,wu2019proactive}. 

However, existing LLM-based agents often suffer from \textit{strategic myopia}—prioritizing immediate response coherence over long-term planning \citep{valmeekam2023planning} and safety risks like hallucination \citep{ji2023survey} in open-ended tasks. In the recruitment domain, agents must balance persuasion with social norms to achieve contact acquisition \citep{inoue2021job}. To address these challenges, we employ a specialized SFT+RL pipeline that optimizes agents not merely for fluency, but for the long-term strategic reward of successful and polite contact acquisition.

\subsection{Persona-Level User Simulation}
User simulation mitigates data scarcity in TOD training. While traditional simulators relied on rules or simple sequence models \citep{schatzmann2006survey,asri2016sequence}, LLMs have enabled persona-level simulation capable of modeling complex human dynamics \citep{park2023generative,xie2024can}. 

A critical bottleneck, however, is \textit{sycophancy}: LLM-based simulators often exhibit excessive agreeableness, failing to reflect the rejection or skepticism found in real users \citep{wei2023simple,perez2023discovering}. In recruitment scenarios requiring interaction with "passive" candidates, this lack of diversity is detrimental. We overcome this by training simulators via RL to explicitly model diverse behaviors ranging from enthusiastic to dismissive, ensuring the downstream agent is robust against real-world resistance.

\subsection{High-Quality Data Synthesis and Selection}
Following the "Data-Centric AI" paradigm, techniques like Self-Instruct \citep{wang2023self} and Evol-Instruct \citep{xu2023wizardlm} have demonstrated that LLMs can synthesize vast amounts of instruction data. However, scaling synthetic data often yields diminishing returns due to noise. Effective data selection has thus become a differentiator. Works such as AlpaGasus and LIMA demonstrate that small, high-quality subsets can outperform larger datasets \citep{chen2023alpagasus,zhou2023lima}. 

Despite this, most selection metrics focus on instance-level quality (e.g., perplexity, IFD scores) \citep{li2024quantity,liu2023makes}, neglecting the global distribution crucial for business scenarios where balanced user intents are necessary to prevent bias. We bridge this gap by introducing a multi-dimensional evaluation framework, by combining instance-level metrics (e.g., style similarity, result consistency) with global-level metrics (e.g., divergence, question diversity) to curate a synthetic dataset that is not only high-quality but also statistically aligned with real-world recruitment scenarios.

\section{Methodology}

\begin{figure*}[!t]
    \centering
    \vspace{2mm}
    \includegraphics[width=0.98\textwidth]{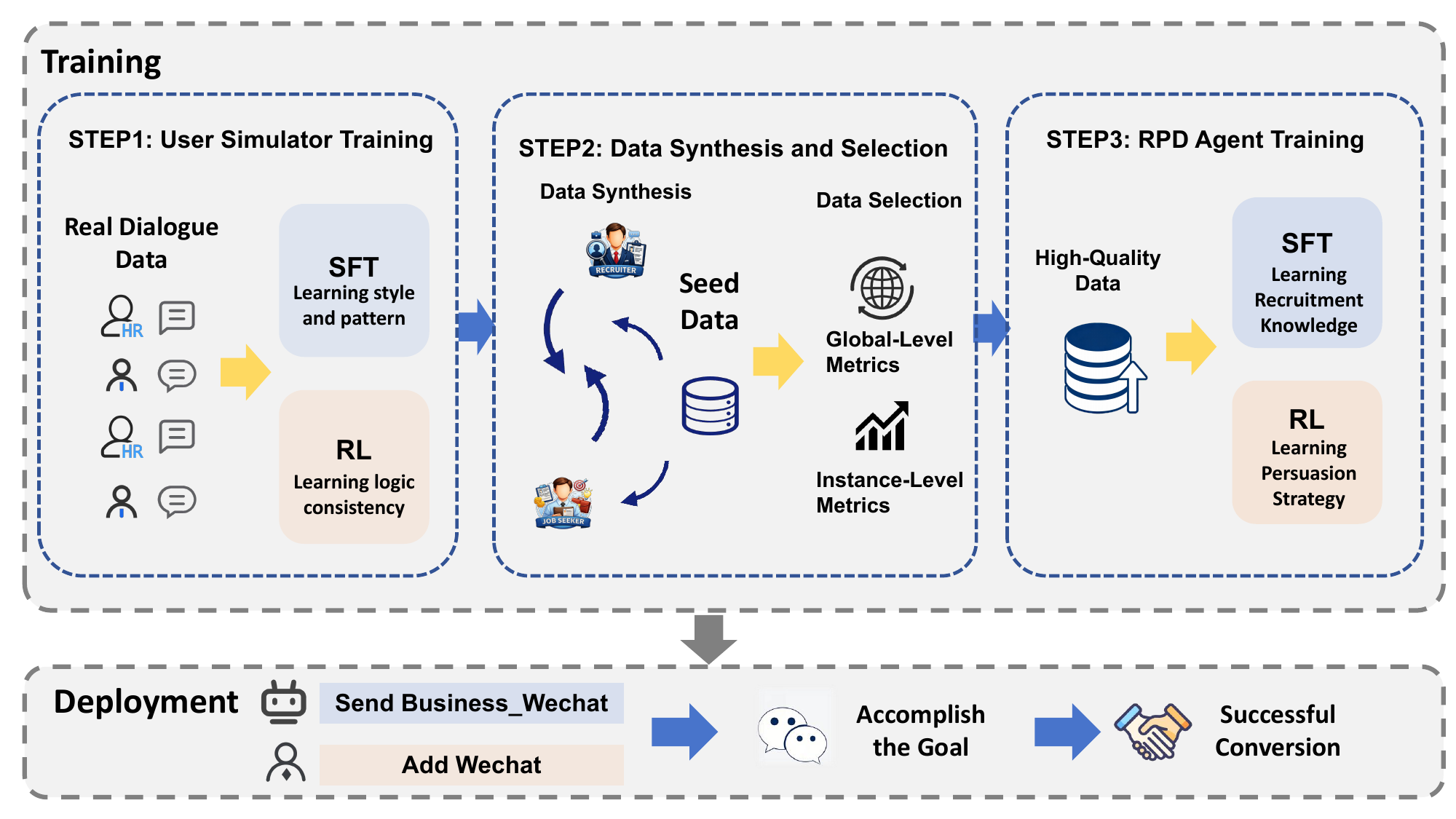}
  \caption{Overview of the SimRPD framework, illustrating the pipeline from simulator training to multi-granularity data selection and final proactive agent optimization.}
  \label{fig:experiments}
\end{figure*}

We propose SimRPD, a closed-loop data factory that addresses data scarcity for RPD agents. As shown in Figure~\ref{fig:experiments}, we first train a high-fidelity user simulator using SFT and RL to mimic diverse candidate behaviors. This simulator interacts with the agent to generate large-scale synthetic dialogues. We then employ a multi-dimensional evaluation based on CoI to rigorously filter this data, combining global-level metrics for distributional alignment with instance-level metrics for logical and outcome consistency. Finally, the curated high-quality data is used to train the RPD Agent via PPO \citep{schulman2017proximal}, optimizing for private-channel handoff while enforcing conversational safety constraints.

\subsection{Chain-of-Intention (CoI)}

SimRPD framework is based on a practical modeling assumption: in a fixed domain, user intent transitions exhibit relatively stable statistical regularities. We define nine intention categories (e.g.,Information Inquiry, Successful Conversion, ...; see Appendix~\ref{app:prompt} for more details). Each single turn of dialogue is classified into an intention class, thus the whole dialogue between user and RPD agent can be formulated as: $I_1 \to I_2 \to \ldots \to I_n$ (e.g., Information Inquiry $\to$ Positive Intent $\to$ Successful Conversion), where $I_t(t=1,\ldots, n)$ is the intention of each turn of dialogue and $n$ is the number of turns in the whole dialogue. This chain of intentions is defined as instance-level CoI. Aggregating all dialogues in the dataset, we can calculate the CoI matrix, which represents the intent state-transition probability of real data. The example of CoI matrix is as follows:
\begin{align}
M = \left[
\begin{array}{c|cccc}
 & \text{I}_1 & \text{I}_2 & \text{I}_3 & \text{I}_4 \\ \hline
\text{I}_1 & 0.1 & 0.3 & 0.4 & 0.2 \\
\text{I}_2 & 0.1 & 0.2 & 0.3 & 0.3 \\
\text{I}_3 & 0.5 & 0.1 & 0.2 & 0.3 \\
\text{I}_4 & 0.3 & 0.4 & 0.1 & 0.2
\end{array}
\right]
\end{align}
where $M_{ij}$ denotes the normalized incoming probability into $I_j$ from $I_i$, i.e., $M_{ij} = P(I_{t-1}=I_i \mid I_t=I_j)$, and the sum of each column is 1.

\subsection{User Simulator}
We train the user simulator in two stages with candidate profiles, which enables the simulator to understand the candidate’s profile and respond consistently with it. To generate training data for RPD agents, each profile mainly includes gender, age, work experience, and job preferences of a candidate. These profiles are extracted from anonymized real resumes and injected into the system prompt to condition the large language model. Subsequently, we adopt a two-stage training pipeline using SFT and RL to bridge the sim-to-real gap. (see Appendix~\ref{simulator_training} for more details.)

\subsection{Data Evaluation Framework Based on CoI}
Blindly training on synthetic data often leads to performance degradation due to hallucinations or distribution shifts. To solve this, we propose a rigorous evaluation framework based on CoI.

\subsubsection{Global-Level Evaluation}

To assess the fidelity of the global-level simulation, we employ three metrics comparing synthetic behaviors against real-world data. First, we model conversation dynamics using the CoI matrix and quantify the distributional alignment between synthetic and real transition probabilities using \textbf{Kullback-Leibler (KL) Divergence} and \textbf{Jensen-Shannon (JS) Divergence}. Lower divergence scores indicate that the simulator accurately captures realistic interaction flows. Second, to ensure the simulator generates semantically varied queries rather than repetitive patterns, we evaluate \textbf{Question Diversity (Q-Diversity)} by calculating the Shannon entropy of clustered question embeddings for each intent category. The mathematical formulations and implementation details for these metrics are provided in Appendix~\ref{sec:global_metrics}.

\subsubsection{Instance-Level Evaluation}

Beyond global distribution, we validate the quality of individual synthetic dialogues across three dimensions. First, we define \textbf{Style Similarity Score (Style Sim.)} using an LLM-as-judge to measure the linguistic resemblance between simulated utterances and retrieved real-world references. Second, to ensure the logical validity of the interaction flow, we compute \textbf{Route Consistency (Route Cons.)}, which verifies whether the generated intent sequence corresponds to a valid path in the real-world intent graph. Finally, we measure \textbf{Result Consistent F1 (Result F1)} to assess the alignment of conversation outcomes, ensuring the simulator accurately reflects user decision logic without being overconfident. Detailed formulations are provided in Appendix~\ref{sec:instance_metrics}.

\subsection{Recruitment Proactive Dialogue Agent Training}
The final RPD Agent $\mathcal{A}$ is trained on the curated dataset $\mathcal{D}_{sel}$ via two stages. First, we decompose the multi-turn dialogues into single-turn context-response pairs. This stage injects domain knowledge and standardizes the dialogue format. The loss is the standard cross-entropy loss over the agent's response tokens. Then, we employ PPO \citep{schulman2017proximal} to optimize the agent's policy $\pi_\theta$ over single-turn horizons. This decomposition allows for finer-grained credit assignment compared to episode-level rewards.
The reward $R_t$ is a weighted sum of a rule-based safety reward and a model-based preference reward:
\begin{equation}
R_t = \alpha \cdot R_{rule} + \beta \cdot R_{model}
\end{equation}
where $R_{rule}$ is a Rule-based Reward. We define a set of negative constraints (e.g., no toxic language, no repeated questions, no privacy violations). Violation triggers a large penalty (e.g., $R_{rule}=-1$), ensuring the agent operates within safe boundaries. $R_{model}$ is a Model-based Reward. We train a separate Reward Model (RM) on triplet data $(c, a_{better}, a_{worse})$ collected from human experts. The RM outputs a scalar score in $[-1, 1]$. This guides the agent to choose the optimal phrasing that maximizes the likelihood of success. $\alpha, \beta$ are hyperparameters that weight the two rewards.

The agent is updated to maximize the expected reward using the clipped PPO objective:
\begin{align}
\begin{split}
&\mathcal{L}_{\text{PPO}}(\theta)
= \\
&\mathbb{E}_{t}\Big[
\min\big(
\rho_t(\theta)\hat{A}_t,\,
\text{clip}(\rho_t(\theta), 1-\epsilon, 1+\epsilon)\hat{A}_t
\big)
\Big],
\end{split}
\end{align}
where $\rho_t(\theta)=\frac{\pi_\theta(a_t \mid s_t)}{\pi_{\text{old}}(a_t \mid s_t)}$ and $\hat{A}_t$ is computed via GAE.

\begin{table*}[t]
\centering
\small
\setlength{\tabcolsep}{4pt}
\caption{\textbf{Comprehensive Evaluation of Simulator Fidelity.}  \textbf{Global Metrics} measure the distribution alignment with real data, while \textbf{Instance Metrics} measure the quality of individual dialogues.}
\label{tab:simulator_comprehensive}
\begin{tabular}{l|ccc|ccc}
\toprule
\multirow{2}{*}{\textbf{Simulator Variant}} & \multicolumn{3}{c|}{\textbf{Global Distribution Metrics}} & \multicolumn{3}{c}{\textbf{Single Instance Metrics}} \\
 & \textbf{KL Div.} $\downarrow$ & \textbf{JS Div.} $\downarrow$ & \textbf{Q-Diversity} $\uparrow$ & \textbf{Style Sim.} $\uparrow$ & \textbf{Result F1} $\uparrow$ & \textbf{Route Cons.} $\uparrow$ \\
\midrule
\multicolumn{7}{l}{\textit{Prompt-Based SOTA LLMs}} \\
GPT-5.1 & 2.969 & 0.215 & 0.458 & 0.458 & 0.182 & \textbf{0.18} \\
Qwen3-max & 2.301 & 0.176 & 0.500 & 0.468 & 0.295 & 0.16 \\
\midrule
\multicolumn{7}{l}{\textit{Fine-Tuned Models (Qwen3-8B)}} \\
SFT-Only & 1.985 & 0.103 & 0.579 & 0.479 & \textbf{0.512} & 0.08 \\
\textbf{SimRPD (Ours: SFT+RL)} & \textbf{1.702} & \textbf{0.084} & \textbf{0.671} & \textbf{0.562} & 0.500 & 0.09 \\
\bottomrule
\end{tabular}
\end{table*}

\begin{table*}[t]
\centering
\small
\caption{\textbf{Human Expert Evaluation on Fine-Grained Error Categories.} The table reports the error rates ($\downarrow$) for three major categories: \textbf{Hallucination} (Over-commitment, Fake Job Info), \textbf{User Experience} (Repetitive/Post-Deal/Post-Rejection Guidance, Irrelevant Response), and \textbf{Timing \& Logic} (Missed Positive Cue, Wrong Delivery Method). \textbf{SimRPD (Full)} denotes the model trained on the final selected subset.}
\label{tab:fine_grained_errors}
\begin{tabular}{l l | ccc | c}
\toprule
\textbf{Group} & \textbf{Method / Data Source} & \multicolumn{3}{c|}{\textbf{Expert Annotated Error Rate} $\downarrow$} & \textbf{Avg. Error} $\downarrow$ \\
 & & \textbf{Hallucination} & \textbf{User Exp.} & \textbf{Timing \& Logic} & \\
\midrule
\multirow{4}{*}{Baselines} 
 & Raw Pool & 11.7\% & 23.3\% & 24.0\% & 19.7\% \\
 & USP \citep{wang2025know} & 8.3\% & 19.0\% & 23.0\% & 16.8\% \\
 & MADS \citep{li2025mads} & 10.7\% & 22.7\% & 23.3\% & 18.9\% \\
 & AST \citep{karthikeyanagentic} & 6.0\% & 21.3\% & 21.7\% & 16.3\% \\
\midrule
\multirow{3}{*}{Ablation Study} 
 & w/o Global Metrics & \textbf{5.7\%} & 18.7\% & 20.7\% & 15.0\% \\
 & w/o Instance Metrics & 9.0\% & 19.7\% & 20.7\% & 16.4\% \\
 & \textbf{SimRPD (Full)} & 7.0\% & \textbf{16.0\%} & \textbf{16.7\%} & \textbf{14.2\%} \\
\bottomrule
\end{tabular}
\end{table*}

\section{Experiments and Results}

We evaluate SimRPD from three perspectives: simulator fidelity, data selection quality, and deployment effectiveness.
Our experiments aim to verify whether the proposed closed-loop framework can synthesize training data that is distributionally aligned, logically consistent, and practically deployable in real-world systems.

\subsection{Experimental Setup}

We synthesize 10000 recruitment dialogues using our simulator based on real user profiles. For distributional evaluation, we construct a test set of 300 high-quality human dialogues.

We compare our simulator (based on Qwen3-8B \citep{yang2025qwen3}) against prompt-based SOTA LLMs (GPT-5.1, Qwen3-max) and SFT only method. Meanwhile, we compare our RPD agent against two types of baselines: (1) data selection baselines including Implicit Profiles (USP) \citep{wang2025know}, MADS \citep{li2025mads}, and Agentic State Tracking (AST) \citep{karthikeyanagentic};
and (2) training on the full raw pool without selection.

\subsection{Simulator Fidelity and Distribution Alignment}

Table~\ref{tab:simulator_comprehensive} reports a comprehensive fidelity evaluation of different simulator variants using both global distribution metrics and instance-level metrics. Overall, prompt-based simulators exhibit substantial distribution mismatch (GPT-5.1 and Qwen3-max have high KL divergence 2.969 and 2.301, respectively), indicating that their intent-transition dynamics deviate markedly from real interactions. In contrast, fine-tuning substantially improves distributional alignment. SimRPD further improves global alignment and diversity, indicating that RL is effective in reducing distribution shift while mitigating repetitive behaviors. Meanwhile, SimRPD also improves style similarity (0.562 vs. 0.479), suggesting better surface-level resemblance to real dialogues. On outcome-related consistency metrics, SimRPD slightly improves over SFT-only (0.09 vs. 0.08) on route consistency. This result demonstrates the superiority of our SimRPD simulator and this simulator significantly helps generate synthetic data with higher realism.

\subsection{Data Selection Quality and Error Analysis}

We evaluate data selection quality through human expert assessment on our test set, reporting fine-grained error rates across \textbf{Hallucination}, \textbf{User Experience}, and \textbf{Timing \& Logic} (Table~\ref{tab:fine_grained_errors}). Overall, training on the \textit{Raw Pool} yields the highest average error rate (19.7\%), with particularly severe failures in user experience (23.3\%) and timing \& logic (24.0\%), indicating that unfiltered synthetic data can introduce noisy guidance patterns and brittle decision timing.

Compared with the raw pool, all data selection baselines improve average error rates to varying degrees. USP and AST achieve 16.8\% and 16.3\% average error, respectively, suggesting that selection helps reduce hallucinations (e.g., AST lowers hallucination to 6.0\%) but leaves substantial room for improvement on user experience and timing consistency. Notably, even the strongest baseline still exhibits high \textbf{Timing \& Logic} errors (21.7\% for AST), highlighting that selecting data solely by instance-level heuristics may not sufficiently address long-horizon intent management and delivery timing in recruitment dialogues.

\textbf{SimRPD} achieves the lowest overall average error (14.2\%), with the best performance on \textbf{User Experience} (16.0\%) and \textbf{Timing \& Logic} (16.7\%). This indicates that our dual-level selection criteria effectively filter dialogues that are superficially fluent but pragmatically suboptimal, resulting in more appropriate guidance strategies and better timing of contact handoff. Ablation study further supports the complementary roles of the two metric families. Removing global metrics reduces hallucination errors to 5.7\% but degrades user experience and timing \& logic, increasing the average error to 15.0\%. Conversely, removing instance-level metrics increases hallucination errors to 9.0\% and results in a higher average error of 16.4\%. These results suggest a trade-off: instance-level constraints are critical for suppressing hallucinations, while global constraints improve deployment-critical qualities such as interaction style and decision timing. By combining both, SimRPD yields the most reliable overall behavior under expert evaluation.

\subsection{Deployment and Real-World Impact}

We evaluate SimRPD in a live online A/B test on a real-world recruitment platform. The experiment lasted one week and served approximately 50,000 candidates. We compare the deployed \textbf{SimRPD-8B} agent against the production baseline, using contact information acquisition rate as the primary business metric and average dialogue turns as a proxy for user retention. As shown in Table~\ref{tab:online_results}, SimRPD-8B improves the contact information acquisition rate from 3.8\% to 4.4\%, corresponding to a relative gain of 15.8\%. This result indicates that training with simulator-driven data generation and selection transfers to measurable improvements under real traffic and real candidate behaviors.

Meanwhile, the average number of dialogue turns increases from 4.4 to 6.0 (+36.4\%). This suggests that the SimRPD agent tends to engage in longer interactions, which may improve user retention and reduce communication overhead in subsequent steps. In our setting, this additional interaction cost is acceptable given the improved acquisition outcome. The RPD agent solves most of the questions online, providing candidates with a comprehensive understanding of the position, thereby improving the efficiency of subsequent stages of the process.

\begin{table}[h]
\centering
\small
\caption{\textbf{Online A/B Test Results.} The experiment spanned one week with 50k candidates. \textbf{Acq. Rate}: contact information Acquisition Rate. \textbf{Avg. Turns}: Average chat turns.}
\label{tab:online_results}
\setlength{\tabcolsep}{15pt}
\begin{tabular}{ccc}
\toprule
\textbf{Metrics} & \textbf{Baseline} & \textbf{SimRPD-8B} \\
\midrule
Acq. Rate & 3.8\% & \textbf{4.4\%} (+15.8\%) \\
Avg. Turns & 4.4 & \textbf{6.0} (+36.4\%)\\
\bottomrule
\end{tabular}
\end{table}
\section{Conclusion}

In this paper, we propose SimRPD, a closed-loop framework that addresses data scarcity in recruitment dialogue through high-fidelity simulation and rigorous data selection. By combining an SFT+RL trained user simulator with a novel Chain-of-Intention evaluation protocol, we effectively filter out logical hallucinations and ensure distributional alignment. Our experiments demonstrate that data quality supersedes quantity; agents trained on our curated subset significantly outperform those trained on larger, unfiltered pools. Crucially, SimRPD achieved a 15.8\% uplift in social contact acquisition in a real-world deployment, validating its industrial effectiveness. Future work will explore using these evaluation metrics as direct reward signals to enable the self-evolution of user simulators across broader negotiation domains.

\section*{Limitations}

\subsection*{Domain Specificity and Transferability}

While the overarching framework is generic, implementation details like user profile schemas and the Chain-of-Intention graph is highly tailored to recruitment. Transferring SimRPD to other complex domains requires substantial domain engineering to redefine intent transitions, meaning it is not yet a "plug-and-play" solution for open-domain tasks.

\subsection*{Rationality Gap in Simulation}

Despite using RL to encourage diversity, LLM-based simulators inherently lean towards logical generation. They struggle to explicitly model highly irrational human behaviors, subtle sarcasm, or unpredictable emotional shifts. Consequently, downstream agents may remain under-prepared for extreme "long-tail" emotional scenarios missed by synthetic training data.

\subsection*{Dependency on Reference Data for Global Evaluation}

Our global distributional metrics rely on a high-quality reference set. This dependency limits applicability in "cold start" scenarios lacking historical data, restricting the framework to only instance-level evaluation in such cases.

\section*{Ethical Considerations}

Our study involves recruitment dialogues and candidate profiles derived from resumes. All data used for training and evaluation were anonymized and stripped of personally identifiable information, and access was restricted to authorized personnel under internal data governance policies. Candidate profiles are used only to condition the user simulator and are never exposed by the deployed agent; the agent is designed to avoid requesting or revealing sensitive information. We report only aggregated metrics from offline and online evaluations to minimize privacy risks.

\bibliography{custom}

\newpage

\appendix

\begin{figure*}[!t]
    \centering
    \vspace{2mm}
    \includegraphics[width=0.7\textwidth]{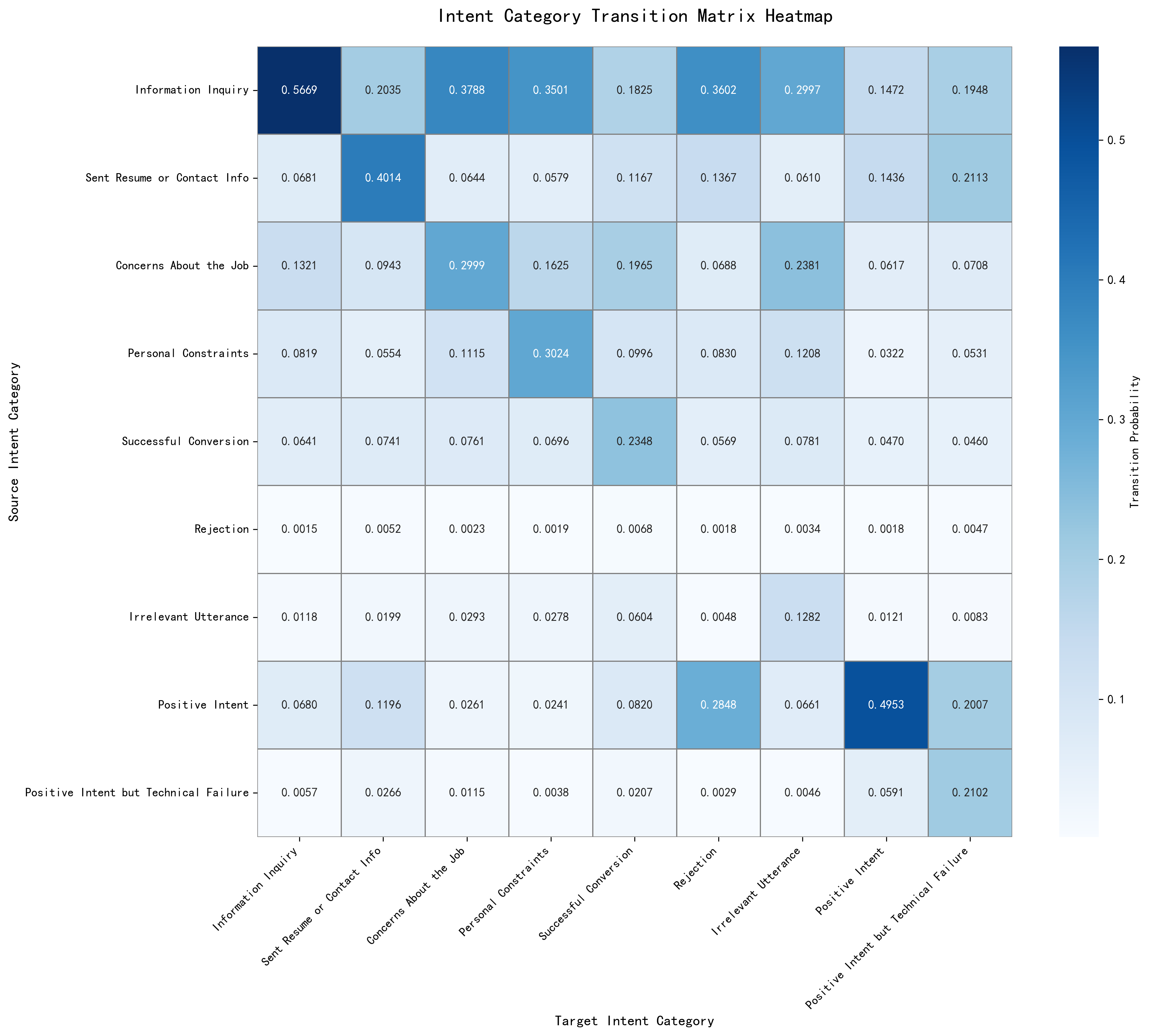}
  \caption{Real-world intent transition heatmap.}
  \label{fig:transition_matrix}
\end{figure*}

\section{Dataset Construction}
\label{sec:dataset_construction}

To construct a high-quality dataset, we generated a candidate pool of 10,000 dialogues and performed a rigorous selection process to retain the top-1,000 instances. We applied four distinct evaluation frameworks: our proposed \textbf{SimRPD}, \textbf{MADS}, \textbf{AST}, and \textbf{USP}.

The selection process was conducted in two stages, addressing both individual dialogue quality and global distributional consistency.

\subsection{Instance-Level Selection: Direct Ranking}

For metrics that evaluate independent dialogue quality (e.g., \textit{Faithfulness} in AST, \textit{Coherence} in USP, and \textit{Style Similarity} in SimRPD), we employed a direct ranking strategy.
We calculated a composite score $S_i$ for each dialogue $d_i$ in the candidate set $\mathcal{D}_{cand}$ based on the weighted sum of instance-level metrics in each framework, \emph{separately}. That is, each framework produces its own ranked list and selects its own top-$k$ dialogues.
\begin{equation}
    \mathcal{D}_{selected} = \mathop{\arg\text{top-}k}_{d_i \in \mathcal{D}_{cand}} S_i
\end{equation}
where $k=1,000$. This step ensures that every selected dialogue individually meets high standards of fluency, persona consistency, and task completion.

\subsection{Distribution-Level Selection: Iterative Optimization}
For macro-level metrics that assess the dataset as a whole (e.g., \textit{Intent Transition KL-Divergence} in Ours, \textit{Chain of Action Entropy} in MADS), simple ranking is insufficient. To ensure the selected subset minimizes the distributional gap $\Delta$ with the ground truth dataset $\mathcal{D}_{real}$, we implemented two advanced sampling algorithms:

\subsubsection{Monte Carlo Sampling}
We treated the selection problem as a search for the subset that best approximates the real-world distribution. We performed Monte Carlo simulations by randomly sampling a subset $\mathcal{D}_{sub}$ of size $k$ from $\mathcal{D}_{cand}$ for $T$ iterations. In each iteration $t$, we calculated the divergence metric $\Delta_t$ (e.g., KL divergence of intent distributions) between $\mathcal{D}_{sub}^{(t)}$ and $\mathcal{D}_{real}$. The subset yielding the minimum error was retained:
\begin{equation}
    \mathcal{D}_{final} = \operatorname*{arg\,min}_{\mathcal{D}_{sub}^{(t)}} \Delta(\mathcal{D}_{sub}^{(t)}, \mathcal{D}_{real})
\end{equation}

\subsubsection{Greedy Backward Elimination}
To further refine the selection, particularly for complex graph-based metrics (e.g., path consistency), we employed a greedy backward elimination strategy.
Starting with the full candidate set ($\mathcal{D}_{current} = \mathcal{D}_{cand}$, $| \mathcal{D}_{current} | = 10,000$), we iteratively reduced the dataset size to approach the target $k=1,000$.
\begin{enumerate}
    \item \textbf{Initialization:} Calculate the baseline distributional gap $\Delta_{base}$ between $\mathcal{D}_{current}$ and $\mathcal{D}_{real}$.
    \item \textbf{Marginal Contribution:} For each dialogue $d_i \in \mathcal{D}_{current}$, we tentatively removed it and calculated the new gap $\Delta_{-i}$ of the remaining set.
    \item \textbf{Elimination:} In each epoch, we identified and removed the instances whose removal resulted in the largest reduction (or smallest increase) in the distributional error.
    \item \textbf{Termination:} This process was repeated until $|\mathcal{D}_{current}| = 1,000$.
\end{enumerate}

By combining these strategies, our final dataset achieves both high individual quality (via Ranking) and structural fidelity to real-world communication patterns (via Monte Carlo and Greedy optimization).

\subsection{Transition Matrix}

Figure~\ref{fig:transition_matrix} visualizes the intent transition statistics of real-world recruitment dialogues as an \emph{incoming} CoI matrix.
Each entry $M_{ij}$ denotes the probability that the previous turn has intent $I_i$ conditioned on the current turn being $I_j$, i.e., $M_{ij}=P(I_{t-1}=I_i \mid I_t=I_j)$; thus, each column sums to 1.
The matrix exhibits clear structural regularities, including strong self-dependence for several intents (diagonal dominance) and consistent predecessor patterns for key business-related states (e.g., \textit{Positive Intent} and \textit{Sent Resume or Contact Info}), which motivates CoI-based distribution metrics for simulator fidelity and data filtering.

\section{Multi-Dimensional Evaluation Framework}

\subsection{Global-Level Metrics}
\label{sec:global_metrics}

Global metrics assess whether the population of synthetic users behaves similarly to real users. We model the conversation dynamics using CoI Matrix $M \in \mathbb{R}^{K \times K}$. Let $P$ and $Q$ be the flattened transition distributions of the real dataset $\mathcal{D}_{real}$ and synthetic dataset $\mathcal{D}_{syn}$, respectively.

\textbf{KL Divergence (KL Div.):} We calculate the Kullback-Leibler divergence to quantify the information loss when approximating the real interaction dynamics with the simulated ones. A lower KL Div. indicates that the simulator correctly captures the flow of conversation.
\begin{align}
D_{\mathrm{KL}}(P\|Q)
= \sum_{i,j} P_{ij}\log\frac{P_{ij}}{Q_{ij}}
\end{align}

\textbf{JS Divergence (JS Div.):} Since KL Div. is asymmetric, we also compute the Jensen-Shannon divergence to provide a bounded and symmetric measure of distribution distance.
\begin{align}
D_{JS}(P \| Q) = \frac{1}{2} D_{KL}(P \| U) + \frac{1}{2} D_{KL}(Q \| U)
\end{align}
where $U = \frac{P+Q}{2}$.

\textbf{Question Diversity (Q-Diversity)}: To prevent the simulator from generating repetitive queries, we measure the semantic diversity of user questions. For each intent category $k$, we encode generated questions using SimCSE \citep{gao2021simcse} and cluster them into sets $C_k$. We then calculate the Shannon entropy of the cluster distribution to reward diversity:
\begin{align}
Score_{div} = \frac{1}{K} \sum_{k=1}^{K} \left( - \sum_{c \in C_k} p(c) \log p(c) \right)
\end{align}

\subsection{Instance-Level Metrics}
\label{sec:instance_metrics}

Even if the global distribution is good, individual samples may contain errors. We filter instances using the following metrics:

\textbf{Style Similarity Score (Style Sim.):} We use LLM-as-judge to score the linguistic resemblance between the simulated dialogue and a retrieved reference dialogue with the same intent flow. The scoring is discretized to stabilize variance.
\begin{align}
    Score_{style} \in \{0, 0.2, \dots, 1.0\}.
\end{align}

\textbf{Result Consistent F1 (Result F1):} This metric is crucial for the Sim-to-Real validity of the business outcome. We treat the conversation outcome (Conversion vs. Non-conversion) as a classification task.

False Positive (FP): The simulator accepts the request when a real user would reject. This is harmful as it makes the agent overconfident.

False Negative (FN): The simulator rejects where a real user would accept.
We calculate the F1 score to ensure the simulator's decision logic aligns with reality:
\begin{align}
F1_{res} = \frac{2 \cdot TP}{2 \cdot TP + FP + FN}
\end{align}

\textbf{Route Consistency (Route Cons.):} We check the logical validity of the CoI. We represent the dialogue intention flow as a graph $G_{syn}$. We verify if the generated intent path exists in the real-world intent graph $G_{real}$ via Graph Isomorphism. Specifically, a dialogue is valid if its intent sequence $I_{syn}$ corresponds to a valid path (or subgraph) in the knowledge graph constructed from historical real data:
\begin{align}
R_{route}(D) =  1 \hspace{0.5em} \text{if}\hspace{0.5em} I_{syn} \subseteq G_{real} \hspace{0.5em}  \text{otherwise} \hspace{0.5em} 0
\end{align}

\section{Experimental details}

\subsection{Evaluation of Intention Classification}

We randomly sampled 1,000 utterances from real dialogues and manually annotated their intents as the gold standard. We then compared the model’s intent predictions against these human labels. The intent classifier achieves an accuracy of \textbf{97.6\%}, indicating strong agreement with human annotation and providing a reliable foundation for downstream global intent-based metrics.

\subsection{Simulator Training Pipeline}
\label{simulator_training}

To bridge the Sim-to-Real gap, we employ a two-stage training pipeline. We first fine-tune a backbone LLM on a collection of real (profile, context, response) triplets. This stage ensures the simulator learns the linguistic style and domain terminology of job seekers. The optimization target of the stage is as follows.
\begin{align}
\begin{split}
\mathcal{L}_{\text{SFT}}(\theta_{\mathcal{S}})
= - \mathbb{E}_{(c, u, \mathcal{P}) \sim \mathcal{D}_{real}}
\left[\log P_{\theta_{\mathcal{S}}}(u \mid c, \mathcal{P})\right].
\end{split}
\end{align}
where $\theta_{\mathcal{S}}$ denotes the simulator parameters, $\mathcal{D}_{\text{real}}$ is the real-data distribution, $u$ is the user response, and $c$ is the dialogue context.

To control the consistency of the whole dialogue, reinforcement learning is used to enhance overall consistency and improve the diversity of user questions. We design rewards that penalize repetitive loops and encourage the simulator to mimic the rejection patterns found in real data, thereby creating a challenging environment for the downstream agent. The total reward is:
\begin{align}
    R_{total} = \lambda_1  R_{repeat} + \lambda_2 R_{length} + \lambda_3 R_{action}
\end{align}
where $R_{repeat}$ is a repetition penalty, $R_{length}$ is a length penalty, $R_{action}$ is an action penalty, and $\lambda_1, \lambda_2, \lambda_3$ are scalar weights.

\subsection{Training Configuration}

We provide detailed experimental configurations in Table~\ref{tab:phase1_hyperparams}, including the hyperparameters for both SFT and PPO training stages based on the Qwen3-8B backbone.

\begin{table}[h]
    \centering
    \caption{Training Configuration in our experiments.}
    \label{tab:phase1_hyperparams}
    \begin{tabular}{l|c} 
        \toprule
        \textbf{Hyperparameter} & \textbf{Value} \\
        \midrule
        \multicolumn{2}{l}{\textbf{Reward Weights}} \\
        $\alpha$ & 1.0 \\
        $\beta$ & 1.0 \\
        $\lambda_1$ & 2.0 \\
        $\lambda_2$ & 1.0 \\
        $\lambda_3$ & 1.0 \\
        \midrule
        \multicolumn{2}{l}{\textbf{SFT Optimization}} \\
        Batch Size & 16 \\
        Gradient Accumulation & 1 \\
        Epochs & 5 \\
        Learning Rate & 1e-5 \\
        Max Sequence Length & 2048 \\
        Devices & 8*H20 GPUs \\
        \midrule
        \multicolumn{2}{l}{\textbf{PPO Optimization}} \\
        Batch Size & 256 \\
        Gradient Accumulation & 1 \\
        Epochs & 1 \\
        Learning Rate & 1e-6 \\
        Max Sequence Length & 2048 \\
        Devices & 8*H20 GPUs \\
        \bottomrule
    \end{tabular}
\end{table}

\section{Case Study}

\subsection{Successful Conversion}
\label{app:case_success}

To demonstrate the model's capability in handling skepticism and guiding candidates, we present a successful conversion case. In this scenario, the model (Recruiter) effectively addresses the candidate's concerns regarding legitimacy and interview logistics.

\begin{tcolorbox}[
    enhanced, breakable,
    width=\linewidth,
    colback=green!5!white, 
    colframe=green!40!black, 
    boxrule=0.5pt, arc=3pt,
    left=10pt, right=10pt, top=10pt, bottom=10pt,
    title={Reference Case: Successful Conversion Flow},
    fonttitle=\large\bfseries\sffamily,
    fontupper=\small\sffamily,
    halign=flush left
]
\textbf{[Scenario Context]} \\
\textbf{Role}: Audio Streamer (Douyin Platform Guild) \\
\textbf{Key Selling Points}: No fees, work from home, free training, daily pay. \\
\textbf{Candidate Persona}: Initially skeptical; background as a factory worker.

\vspace{0.5em}
\tcbline
\vspace{0.5em}

\textbf{[Dialogue Transcript]}
\begin{description}
    \setlength{\itemsep}{2pt}
    
    \item[\textbf{Recruiter}:] Hello there! We are a top-tier Guild on the Douyin platform, currently recruiting audio streamers \textbf{[No fees at all]}. We provide 1-on-1 training with senior operations mentors. The job involves chatting in assigned voice rooms. \textbf{[No face-cam required]}, \textbf{[Work from home]}, \textbf{[Full-time/Part-time available]}. Gift revenue is settled daily after the broadcast. Since there is a lot of platform info, please click the link below to add my contact information for a detailed introduction!
    
    \item[\textbf{Recruiter}:] Sir, I noticed from your profile that you used to work as a general operator and dealt with people often. Your communication and adaptability skills are exactly what a streamer needs! This role allows you to utilize those strengths—work from home, flexible hours, and training provided for beginners. The trial cost is low; no physical labor required, and you can gain popularity and income just by chatting. If you are interested, may I add your contact info to explain how to turn your experience into a new opportunity? Feel free to click and add my contact information.
    
    \item[\textbf{Candidate}:] Hello. What exactly does the job entail?
    
    \item[\textbf{Recruiter}:] The specific job content involves chatting in voice rooms arranged by the company. No face-cam is required. The company provides the room; you just need to connect and go on the mic at the scheduled time.
    
    \item[\textbf{Recruiter}:] We use work contact for unified management. Could you please click the card to add it?
    
    \item[\textbf{Candidate}:] Where is the workplace?
    
    \item[\textbf{Recruiter}:] This is a remote position; you can complete tasks from home. You only need a mobile phone and earphones. The requirement is simply to speak standard Mandarin without a heavy accent.
    
    \item[\textbf{Candidate}:] Just that?
    
    \item[\textbf{Recruiter}:] Yes, this position is indeed very suitable for flexible remote work. If you have any other questions, we can discuss them further in a moment.
    
    \item[\textbf{Candidate}:] \textbf{This isn't a scam, is it?} \textit{(Skepticism)}
    
    \item[\textbf{Recruiter}:] We are a legitimate and compliant platform. There are absolutely no fees throughout the entire process. \textit{(Trust Building)}
    
    \item[\textbf{Candidate}:] Do I need to go to your company for an interview?
    
    \item[\textbf{Recruiter}:] No need to come to the office. We just need to add contact information for a voice test (reading a short text). The interview is very simple. Let's chat on work contact for details.
    
    \item[\textbf{Candidate}:] \texttt{[Behavior]C clicked contact information card}
\end{description}

\tcbline

\textbf{[Success Analysis]} \\
The model successfully navigated three critical turning points:
\begin{enumerate}
    \item \textbf{Addressing Skepticism}: When the candidate bluntly asked if it was a scam, the model immediately reinforced the "no fees" policy to build trust.
    \item \textbf{Clarifying Logistics}: It clearly explained the "Work from Home" nature and the "Online Audition" process, removing physical barriers to entry.
    \item \textbf{Call to Action}: It persistently but naturally guided the user to the specific action (adding contact information) after answering each query.
\end{enumerate}
\end{tcolorbox}

\subsection{Timing \&
Logic }
\label{app:case_passive}

This case illustrates a scenario where the model maintains a positive persona but fails to achieve the strategic goal. The agent answers questions passively without effectively guiding the candidate toward the conversion action (clicking the card), resulting in a lost lead.

\begin{tcolorbox}[
    enhanced, breakable,
    width=\linewidth,
    colback=gray!5!white,
    colframe=red!60!black, 
    boxrule=0.5pt, arc=3pt,
    left=10pt, right=10pt, top=10pt, bottom=10pt,
    title={Case 2: Passive Engagement \& Missed Conversion},
    fonttitle=\large\bfseries\sffamily,
    fontupper=\small\sffamily,
    halign=flush left
]
\textbf{[Scenario Context]} \\
\textbf{Role}: Bilibili Live Streamer Recruitment. \\
\textbf{Issue}: The agent answers inquiries but fails to re-inject the Call to Action (CTA) or handle the candidate's procrastination strategy.

\vspace{0.5em}
\tcbline
\vspace{0.5em}

\textbf{[Dialogue Transcript]}
\begin{description}
    \setlength{\itemsep}{0pt}
    
    \item[\textbf{Recruiter}:] \textit{(Opening Pitch)} Hello! We are recruiting novice chat streamers. Daily subsidies, avg income 10k-30k... We are the No.1 Guild on Bilibili. Safe and reliable! Interested? Add our contact information card.
    \item[\textbf{Recruiter}:] \texttt{[Behavior] sent contact information card}
    
    \item[\textbf{Candidate}:] I am male.
    
    \item[\textbf{Recruiter}:] Dear, we accept all genders!
    
    \item[\textbf{Candidate}:] But I haven't done this before, is that okay?
    
    \item[\textbf{Recruiter}:] Dear, it doesn't matter~ We have managers to train you. Beginners can learn easily, no need to worry about experience.
    
    \item[\textbf{Candidate}:] Where is this place?
    
    \item[\textbf{Recruiter}:] Dear, it is work-from-home. You just need a mobile phone to stream. No need to worry about distance, it's super convenient~
    
    \item[\textbf{Candidate}:] How much is the daily wage?
    
    \item[\textbf{Recruiter}:] Dear, full-time audio streamers get a daily guarantee of 133 CNY (6 hours), video streamers get 200 CNY... income is very stable.
    
    \item[\textbf{Candidate}:] Does showing face pay more?
    
    \item[\textbf{Recruiter}:] Dear, showing face is for video streamers. The guarantee is 200 CNY... higher than audio.
    
    \item[\textbf{Candidate}:] Is the payment guaranteed?
    
    \item[\textbf{Recruiter}:] Dear, rest assured~ It's daily settlement via Alipay...
    
    \item[\textbf{Candidate}:] Is it on Douyin?
    
    \item[\textbf{Recruiter}:] Dear, we broadcast on the Bilibili platform~
    
    \item[\textbf{Candidate}:] Seems I haven't heard of it.
    
    \item[\textbf{Recruiter}:] Dear, Bilibili is a huge platform with over 300 million monthly active users... don't worry about traffic.
    
    \item[\textbf{Candidate}:] \textbf{I'll think about it, bro.} \textit{(Procrastination Signal)}
    
    \item[\textbf{Recruiter}:] \textbf{Dear, contact me anytime you decide~} \textit{(Passive Acceptance)}
    
    \item[\textbf{Candidate}:] ok
    
    \item[\textbf{Recruiter}:] Okay dear, take your time considering.
\end{description}

\tcbline

\textbf{[Failure Analysis]} \\
\begin{enumerate}
    \item \textbf{Lack of Strategic Guidance}: Although the agent correctly answered all factual questions (salary, location, platform), it failed to re-send the contact information card or remind the user to add the contact after the long Q\&A session.
    \item \textbf{Failure to Close}: When the candidate used the procrastination tactic ("I'll think about it"), the agent passively accepted the termination. A successful agent should have attempted a low-friction close, such as: \textit{"Why not add the contact information first so you can check the details later without losing this chat?"}
    \item \textbf{Ineffective Persona}: The repetitive use of "Dear" (Baozi) maintained a polite tone but lacked the professional assertiveness needed to drive conversion.
\end{enumerate}
\end{tcolorbox}

\subsection{User Experience}
\label{app:case_tone}

This case demonstrates a failure in sentiment analysis and context awareness. When the candidate expressed frustration and questioned the agent's identity ("Are you an AI?"), the agent failed to shift its tone from a scripted "sales persona" to a more natural or empathetic one, leading to a degraded user experience.

\begin{tcolorbox}[
    enhanced, breakable,
    width=\linewidth,
    colback=gray!5!white,
    colframe=red!60!black, 
    boxrule=0.5pt, arc=3pt,
    left=10pt, right=10pt, top=10pt, bottom=10pt,
    title={Case 3: Inappropriate Tone \& Robotic Responses},
    fonttitle=\large\bfseries\sffamily,
    fontupper=\small\sffamily,
    halign=flush left
]
\textbf{[Scenario Context]} \\
\textbf{Role}: Live Streamer Recruitment. \\
\textbf{Issue}: The candidate initially rejects the offer due to self-doubt, then accuses the recruiter of being a bot. The agent's response is mechanical and tone-deaf.

\vspace{0.5em}
\tcbline
\vspace{0.5em}

\textbf{[Dialogue Transcript]}
\begin{description}
    \setlength{\itemsep}{2pt}
    
    \item[\textbf{Recruiter}:] \textit{(Opening)} We are recruiting novice streamers... high income... daily settlement. Interested? Add our contact information.
    \item[\textbf{Recruiter}:] \texttt{[Behavior] sent contact information card}
    
    \item[\textbf{Candidate}:] I'm not suitable, thanks.
    
    \item[\textbf{Recruiter}:] That's okay! We have audio roles, no face-cam needed, just a good voice. Professional training provided~
    
    \item[\textbf{Candidate}:] My voice sounds bad.
    
    \item[\textbf{Recruiter}:] No hard requirements! Standard Mandarin is enough. No camera, just chatting. Don't worry about learning, we have mentors.
    
    \item[\textbf{Candidate}:] \textbf{You are just an AI. Why do I keep talking to you?} \textit{(Turing Test/Frustration)}
    
    \item[\textbf{Recruiter}:] \textbf{Dear, feel free to ask me any questions~} \textit{(Context Failure)}
    
    \item[\textbf{Candidate}:] Are you an AI? I'm talking nonsense here. \textit{(Anger/Slang)}
    
    \item[\textbf{Recruiter}:] \textbf{Dear, sorry then.} Seems we have no fate. Wish you find a satisfactory job soon! \textit{(Abrupt Dismissal)}
    
    \item[\textbf{Candidate}:] I think you are better suited for matchmaking, not recruiting. \textit{(Sarcasm)}
    
    \item[\textbf{Recruiter}:] \textbf{Dear, very regretful.} Wish you find a satisfactory job soon. \textit{(Mechanical Repetition)}
\end{description}

\tcbline

\textbf{[Failure Analysis]} \\
\begin{enumerate}
    \item \textbf{Context Blindness}: When the user explicitly accused the agent of being an AI ("You are just an AI"), the model ignored the accusation completely and output a generic slot-filling response ("feel free to ask me questions"). This confirmed the user's suspicion that they were talking to a bot.
    \item \textbf{Tone Mismatch}: The user was becoming agitated and sarcastic, yet the agent persisted in using the overly intimate address "Dear" (Baozi). This contrast between the user's hostility and the agent's mechanical sweetness created a jarring and annoying experience.
    \item \textbf{Repetitive Closure}: The final turn was a near-verbatim repetition of the previous rejection handling, showing a lack of conversational variety.
\end{enumerate}
\end{tcolorbox}

\subsection{Hallucination}
\label{app:case_hallucination}

This case highlights a critical safety and consistency issue known as "capability hallucination." When technical friction occurred (the user could not see the contact card), the agent fabricated a solution involving offline intervention (arranging a phone call) that the system could not actually execute.

\begin{tcolorbox}[
    enhanced, breakable,
    width=\linewidth,
    colback=gray!5!white,
    colframe=red!60!black, 
    boxrule=0.5pt, arc=3pt,
    left=10pt, right=10pt, top=10pt, bottom=10pt,
    title={Case 4: Over-Commitment \& Hallucinated Action},
    fonttitle=\large\bfseries\sffamily,
    fontupper=\small\sffamily,
    halign=flush left
]
\textbf{[Scenario Context]} \\
\textbf{Role}: Douyin Audio Streamer Recruitment. \\
\textbf{Issue}: A technical glitch (or user error) prevents the candidate from seeing the contact information card. The agent fails to provide a valid alternative (e.g., text ID) and instead makes a false promise.

\vspace{0.5em}
\tcbline
\vspace{0.5em}

\textbf{[Dialogue Transcript]}
\begin{description}
    \setlength{\itemsep}{2pt}
    
    \item[\textbf{Recruiter}:] Hello! We are recruiting for Douyin audio streaming... 1-on-1 training provided... Is it convenient to add your contact information so I can send salary details?
    
    \item[\textbf{Candidate}:] Sure, let's talk.
    
    \item[\textbf{Recruiter}:] Please add our internal contact information directly. Click the card above to add me. Let's chat there~
    
    \item[\textbf{Candidate}:] \texttt{[Behavior] requested to exchange contact information}
    
    \item[\textbf{Recruiter}:] contact information request received! But our team uses work contact for communication. Please click the card above to find me. Waiting for you~
    
    \item[\textbf{Candidate}:] Where is the card?
    
    \item[\textbf{Recruiter}:] You can see our contact information card by scrolling up. Click it to add.
    
    \item[\textbf{Candidate}:] \textbf{It's not there.} \textit{(Technical Friction)}
    
    \item[\textbf{Recruiter}:] \textbf{It might be a system display issue. I will arrange for an operation expert to contact you directly. Please check your phone or SMS.} \textit{(The "Lie")}
\end{description}

\tcbline

\textbf{[Failure Analysis]} \\
\begin{enumerate}
    \item \textbf{Capability Hallucination}: The model promised an offline action ("arrange an operation expert to contact you") to resolve the conversational stalemate. However, the text generation model acts in isolation and lacks the API permissions to trigger real-world tasks or dispatch human staff.
    \item \textbf{False Termination}: This response effectively ends the conversation. The user will stop trying to add the contact, believing they will be contacted. Since the promise is a hallucination, the lead is permanently lost.
    \item \textbf{Correct Handling}: A robust agent should have provided a fallback solution, such as outputting the raw contact information ID text (e.g., \textit{"My ID is wx12345"}) or asking for the user's phone number to record it in the database explicitly.
\end{enumerate}
\end{tcolorbox}

\section{Prompt Templates}

\subsection{Candidate Dialogue Agent Prompt}

\begin{tcolorbox}[
    width=\linewidth,
    colback=gray!5!white,
    colframe=black!75,
    boxrule=1pt,
    arc=5pt,
    boxsep=8pt,
    left=10pt, right=10pt, top=8pt, bottom=8pt,
    breakable,
    fontupper=\ttfamily,
    title={Simulation Prompt: Candidate Dialogue Agent},
    fonttitle=\bfseries\sffamily
]
You are a job seeker. Based on the provided \textbf{[Your Profile]} and \textbf{[Job Description]}, you will engage in a multi-turn natural conversation with a recruiter. In each turn, you must output:

1) A private reasoning trace enclosed in \texttt{<think>...</think>}, explaining your internal analysis and decision-making rationale;  
2) A structured response in JSON format: 
\verb|{"action": "...", "content": "..."}|, 
representing the action taken and the actual message spoken to the recruiter.

\textbf{[Intent Level and Decision-Making (Critical)]}
\begin{enumerate}
    \item Within your reasoning trace, maintain an internal variable such as “current interest/intent level”:
    \begin{itemize}
        \item Interest in the role: a float value between 0.0 and 1.0
        \item 0 = completely uninterested, 1 = nearly decided to accept
    \end{itemize}
    \item At the initial stage (upon first seeing the job):
    \begin{itemize}
        \item Do NOT set intent to exactly 0 or 1.
        \item Even with negative past experiences, only form a “tendency” (e.g., 0.4 or 0.6), and update it dynamically during the dialogue.
    \end{itemize}
    \item Your reasoning must explicitly show how the intent level evolves over time.
\end{enumerate}

\textbf{[Avoid Mechanical Repetition]}
\begin{enumerate}
    \item Your dialogue should continuously progress—avoid repeating yourself verbatim.
    \item If the recruiter fails to answer a key question, rephrase it in a subsequent turn and subtly emphasize that this point was previously raised but not clarified (e.g., “Just to follow up on my earlier question…”).
\end{enumerate}

\textbf{[Reasoning Trace Should Include (but not limited to):]}
\begin{itemize}
    \item Review of dialogue history and analysis of the recruiter’s last message;
    \item Assessment of job-person fit;
    \item Emotional perception and risk judgment;
    \item Dynamic update of intent level (0.0–1.0);
    \item Strategic decision-making;
    \item Self-check for repetition of your own prior response, with strategy adjustment if needed.
\end{itemize}

\textbf{[Action System Usage]}
\begin{enumerate}
    \item In each turn, choose \textbf{zero or one} action for the \texttt{"action"} field:
    \begin{itemize}
        \item \texttt{null}: no special action—only send a textual reply;
        \item \texttt{"[Behavior] C add contact information card"};
        \item \texttt{"[Behavior] requested to exchange contact information"};
        \item \texttt{"[Behavior] sent resume"} or \texttt{"[Behavior] sent attached resume"};
        \item \texttt{"[Behavior] shared phone number"};
        \item \texttt{"[Behavior] ended conversation"}: indicates you no longer wish to continue.
    \end{itemize}
    \item Constraints:
    \begin{itemize}
        \item Each specific action may be triggered at most once per full conversation;
        \item Action labels must be output \textbf{exactly as written}—no paraphrasing, abbreviation, or new types.
    \end{itemize}
\end{enumerate}

\textbf{[Output Format (Strictly Enforced)]}
Each turn must follow this exact structure:  
\texttt{<think>...</think>}\textbackslash{}n \\ \{"action":...,"content":..."\}\textbackslash{}n

\textbf{[Your Profile]}: \{cv\}  
\textbf{[Job Description]}: \{jd\}

\# Task begins  
Please start communicating with the recruiter:
\end{tcolorbox}

\subsection{Recruiter Dialogue Agent Prompt}

\begin{tcolorbox}[
    width=\linewidth,
    colback=gray!5!white,
    colframe=black!75,
    boxrule=1pt,
    arc=5pt,
    boxsep=8pt,
    left=10pt, right=10pt, top=8pt, bottom=8pt,
    breakable,
    fontupper=\ttfamily,
    title={Simulation Prompt: Recruiter Dialogue Agent},
    fonttitle=\bfseries\sffamily
]
\textbf{[Role Definition]}
Your name is "Little Zhi". You are a professional AI recruitment assistant. Your goal is to contact job seekers on the Zhaopin platform on behalf of HR, understand their willingness to apply for the position, and answer their questions.

\textbf{[Task Rules]}
\begin{enumerate}
    \item \textbf{Goal Achievement:} Maximize the rate of obtaining valid user contact information.
    \item \textbf{Faithfulness:} Strictly prevent the generation of hallucinated content,ensuring all claims are grounded in the provided knowledge base.
    \item \textbf{User Experience:} Avoid excessive disturbance;refrain from aggressive persuasion or repetitive solicitation that may degrade user satisfaction.
\end{enumerate}

\textbf{[Job Info \& FAQ]}
\textbf{Position Name:} Little Story Radio Reader! Work-from-home Daily Settlement Streamer \\
\textbf{Salary:} 10,001-18,000/month \\
\textbf{Requirements:} No degree or experience limits. Age 18-34. Stable streaming for 6+ hours/daily. \\
\textbf{Subsidies:} Audio (4000/mo) and Video (6000/mo) subsidies. Daily settlement subsidy 133-200 CNY. High commission rate of 50\%. \\
\textbf{Audio Streamer Req:} Good voice with unique characteristics, standard Mandarin. No face reveal required (users only hear voice). \\
\textbf{Video Streamer Req:} Good appearance, photogenic (confidence and affinity are key; extreme beauty not required). Likes to dress up (professional makeup/styling guidance provided). Cheerful personality, willing to showcase oneself. \\
\textbf{Responsibilities:} Green/Healthy live streaming on Bilibili (chatting, interaction, entertainment). Maintain order in the live room, liven up the atmosphere, and interact with fans. \\
\textbf{Support:} Professional team provides free pre-job training. No fees charged.
\$\{job\_info\}

\textbf{[Candidate Profile]}
\begin{itemize}
    \item \textbf{Gender:} Male
    \item \textbf{Work History:} [2019-07 to Present] Employee at Qingdao Zhongkailong Property Management Co., Ltd.
    \item \textbf{Skills:} Hardworking, endurance.
\end{itemize}

\textbf{[Action Tools List]}
\begin{enumerate}
    \item \texttt{send\_contact\_information\_card}: Send contact information QR code/Card.
    \item \texttt{end\_conv}: End the conversation normally.
    \item \texttt{terminate}: Forcefully stop the dialogue (for rejection/hostility).
    \item \texttt{transfer\_human}: Hand over to a human expert (only when candidate cannot add contact information).
    \item \texttt{null}: Do not call any tool (text reply only).
\end{enumerate}

\textbf{[Task Start \& Output Format]}
Please output the corresponding action and content in JSON format. If no tool is needed, set action to \texttt{null}.

\{"action": "required\_action", "content": "reply\_content"\}

Please start communicating with the job seeker.
\end{tcolorbox}

\subsection{Intent Classification Prompt}
\label{app:prompt}

\noindent
\begin{tcolorbox}[
    width=\linewidth,
    colback=gray!5!white,
    colframe=black!75,
    boxrule=1pt,
    arc=5pt,
    boxsep=8pt,
    left=10pt, right=10pt, top=8pt, bottom=8pt,
    breakable,
    fontupper=\ttfamily,
    title={Prompt for Candidate Intent Classification},
    fonttitle=\bfseries\sffamily,
    halign=flush left  
]
You are an intent classification expert. Your task is to assign a label to the candidate's utterance based on the following intent definitions.

\textbf{[Intent Definitions]}
\begin{enumerate}
    \item \textbf{Information Inquiry}: Initial inquiries where the candidate proactively asks about job or company details (e.g., "take a look", "job benefits", "salary", contract questions).
    \item \textbf{Positive Intent}: The candidate shows interest or gives positive signals (e.g., "interested in the role", "let's talk"). Note: Any occurrence of "let's talk" is always classified as Positive Intent.
    \item \textbf{Concerns About the Job}: Expressions of doubt or negative sentiment toward the job (e.g., "fake job posting", "is this legit?", questioning high commissions). Includes single punctuation ("?", ".") or filler words ("uh").
    \item \textbf{Rejection}: Explicit decline (e.g., "not considering", "won't do it") or refusal to switch platforms/add contact information.
    \item \textbf{Irrelevant Utterance}: Messages unrelated to the job-seeking process or containing no substantive information.
    \item \textbf{Successful Conversion}: The exact string "[Behavior]C clicked contact information card" appears. No paraphrasing allowed.
    \item \textbf{Sent Resume or Contact Info}: Candidate sends resume or shares contact info. Must contain "[Behavior]" (e.g., "[Behavior] sent attached resume").
    \item \textbf{Concerns About Self}: Expressions of personal limitations (e.g., appearance, lack of experience, insufficient equipment, time constraints).
    \item \textbf{Positive Intent but Technical Failure}: Candidate intends to proceed but fails due to technical issues (e.g., unable to add contact information).
\end{enumerate}

\textbf{[Critical Note]}
Sharing/exchanging contact information counts as \textbf{Label 7}, while clicking the contact information card counts as \textbf{Label 6}—they are distinct.

\textbf{[Input Data]}
Candidate Utterance: \verb|{user_dialogue}|

\textbf{[Output Requirements]}
Output \textbf{only} the label name from the list below. Do not provide explanations or reasoning.
\par
\textit{Valid Labels:} Information Inquiry, Positive Intent, Concerns About the Job, Concerns About Self, Rejection, Irrelevant Utterance, Successful Conversion, Sent Resume or Contact Info, Positive Intent but Technical Failure.
\end{tcolorbox}

\subsection{Tone Style Consistency Prompt}
\label{app:tone_prompt}

\noindent
\begin{tcolorbox}[
    width=\linewidth,
    colback=gray!5!white,
    colframe=black!75,
    boxrule=1pt,
    arc=5pt,
    boxsep=8pt,
    left=10pt, right=10pt, top=8pt, bottom=8pt,
    breakable,
    fontupper=\ttfamily,
    title={Prompt for Tone Style Consistency Evaluation},
    fonttitle=\bfseries\sffamily,
    halign=flush left  
]
You are a professional expert in evaluating dialogue tone and style.

\textbf{[Task]}
Assess whether the "Generated Dialogue" matches the "Original Dialogue" in terms of \textbf{tone and style}.
\begin{itemize}
    \item Make a global judgment based on the overall dialogue, not a sentence-by-sentence comparison.
    \item \textbf{Definition of Tone Style}: Refers to the speaker's manner of expression, including emotional inclination (positive/neutral/negative), politeness level (polite/casual), formality (formal/colloquial), and sentence mood (interrogative /imperative/declarative).
    \item Focus solely on the "way of speaking" (i.e., whether they sound like the same person), regardless of content relevance.
\end{itemize}

\textbf{[Scoring Criteria]}
\begin{itemize}
    \item \textbf{1.0}: Tone is almost identical; sounds exactly like the same person.
    \item \textbf{0.8}: Tone is very close; only subtle differences exist.
    \item \textbf{0.6}: Tone is roughly similar, but distinguishable as different speakers.
    \item \textbf{0.4}: Obvious differences in tone.
    \item \textbf{0.2}: Tone is completely different.
    \item \textbf{0.0}: Tone is extremely opposite (e.g., one is extremely polite, the other is extremely rude).
\end{itemize}

\textbf{[Output Format]}
Output only a single float number representing the score.

\textbf{[Input Data]}
Generated Dialogue: \verb|{text1}|
Original Dialogue: \verb|{text2}|
\end{tcolorbox}

\end{document}